\documentclass{article}

    \PassOptionsToPackage{numbers, compress}{natbib}

\usepackage[preprint]{neurips_2024}

\usepackage[utf8]{inputenc} 
\usepackage[T1]{fontenc}    
\usepackage{hyperref}       
\usepackage{url}            
\usepackage{booktabs}       
\usepackage{amsfonts}       
\usepackage{nicefrac}       
\usepackage{microtype}      
\usepackage{xcolor}         
\usepackage{amsmath}
\usepackage{multicol}
\usepackage{multirow}
\usepackage{graphicx}
\usepackage{makecell}

\title{NubbleDrop: A Simple Way to Improve Matching Strategy for Prompted One-Shot Segmentation }

\author{%
  Zhiyu Xu \\
  College of Information Science and Technology\\
  Jinan University\\
  \texttt{xuzhiyu@stu2022.jnu.edu.cn} \\
\And
 Qingliang Chen\thanks{Corresponding author} \\
  Department of Computer Science\\
  Jinan University\\
  \texttt{tpchen@jnu.edu.cn} \\
}

\begin{document}

\maketitle

\begin{abstract}
Driven by large-data-trained segmentation models, such as SAM \cite{kirillov2023segment}, research in one-shot segmentation has experienced significant advancements. Recent contributions like PerSAM \cite{zhang2023personalize} and MATCHER \cite{liu2023matcher}, presented at ICLR 2024, utilize a similar approach by leveraging SAM with one or a few reference images to generate high-quality segmentation masks for target images. Specifically, they utilize raw encoded features to compute cosine similarity between patches within reference and target images along the channel dimension, effectively generating prompt points or boxes for the target images—a technique referred to as the matching strategy. However, relying solely on raw features might introduce biases and lack robustness for such a complex task. To address this concern, we delve into the issues of feature interaction and uneven distribution inherent in raw feature-based matching. In this paper, we propose a simple and training-free method to enhance the validity and robustness of the matching strategy at no additional computational cost (\textbf{NubbleDrop}). The core concept involves randomly dropping feature channels (setting them to zero) during the matching process, thereby preventing models from being influenced by channels containing deceptive information. This technique mimics discarding pathological nubbles, and it can be seamlessly applied to other similarity computing scenarios. 
We conduct a comprehensive set of experiments, considering a wide range of factors, to demonstrate the effectiveness and validity of our proposed method. Our results showcase the significant improvements achieved through this simmple and straightforward approach.
\end{abstract}

\section{Introduction}

Research in Vision Foundation Models (VFMs) has made tremendous strides in recent times. Fueled by extensive image-text contrastive pre-training, CLIP (Radford et al., 2021) \cite{radford2021learning} and ALIGN (Jia et al., 2021) \cite{jia2021scaling} demonstrate robust zero-shot transfer capabilities across a wide array of classification tasks. DINOv2 (Oquab et al., 2023) \cite{oquab2023dinov2} showcases remarkable proficiency in visual feature matching, enabling it to comprehend intricate information at both the image and pixel levels, solely from raw image data. Furthermore, the Segment Anything Model (SAM) (Kirillov et al., 2023) \cite{kirillov2023segment} has achieved impressive class-agnostic segmentation performance by training on the SA-1B dataset, comprising 1 billion masks and 11 million images.

However, unlike Large Language Models (LLMs), which seamlessly integrate various language tasks using a unified model structure and pre-training approach, VFMs face challenges when directly addressing diverse perception tasks. For instance, these approaches often require a task-specific model architecture and fine-tuning for each specific task (He et al., 2022 \cite{he2022masked}; Oquab et al., 2023 \cite{oquab2023dinov2}).

To enhance the transferability of Vision Foundation Models (VFMs), efforts have been made by PerSAM \cite{zhang2023personalize} and Matcher \cite{liu2023matcher} to prompt the VFM, specifically SAM. They employ a systematic approach where each category is prompted with a single reference photo. This process involves several key steps, including matching the target image with the reference images, extracting prompted points or boxes to guide SAM in generating segmentation masks, and utilizing the matched information to select segmentation masks and produce the final results.

 Upon reevaluation of this process, we've identified two significant concerns. Firstly, Vision Foundation Models (VFMs) entail intricate feature interactions and fusion during feature extraction and processing, which may not accurately depict the similarity between patches at the channel level. In essence, not all individual channel-to-channel disparities can effectively capture the similarity between two patches. Secondly, we've observed that certain channels' absolute values within the extracted features from the same patch are excessively large, exerting a dominant influence on the contrast while overshadowing information from numerous other channels.

Addressing the first issue, we conducted experiments to compute the matching degree of background points and target object points in a substantial number of images separately. The findings revealed that for numerous images, the optimal matching points for their target object points were situated within the background, highlighting significant deficiencies in the existing matching strategy. Concerning the second issue, we designed experiments to calculate the extreme values and variances of normalized channel values (absolute values) across a large dataset of images. These experiments demonstrated a highly unbalanced distribution of channel values, where certain channel values played a pivotal role in the matching process. Should these points also encounter the first issue, it could result in misjudgment, thereby significantly impairing the model's performance.


To mitigate these two concerns without incurring substantial computational costs, we have introduced a straightforward method (\textbf{NubbleDrop}) that imposes minimal computational overhead. The crux of this approach involves randomly setting a small subset of feature channels to zero. We have demonstrated that for individual images, NubbleDrop has negligible impact on the model's original performance. However, there exists a probability that dropping channels affected by either of the aforementioned issues can notably enhance the accuracy of image segmentation. Consequently, when applied to large-scale image datasets, this operation yields considerable improvements. Furthermore, even for datasets with a mean Intersection over Union (mIoU) approaching 90, this operation still delivers significant enhancements. Additionally, we discuss alternative operations for enhancing the matching strategy that entail slightly more computational complexity in the \textbf{Methodology} section.

Our comprehensive experiments underscore the superior generalization performance of NubbleDrop across various segmentation tasks. In the realm of one-shot semantic segmentation, Matcher with NubbleDrop (\textbf{MN}) achieves a remarkable 53.5 mIoU on COCO-20$^{i}$ \cite{nguyen2019feature}, surpassing the state-of-the-art specialist model by 2.4 and the original model by 1.4. Similarly, MN achieves a notable 34.0 mIoU on LVIS-92$^{i}$ \cite{liu2023matcher}, outperforming the state-of-the-art generalist model SegGPT (Wang et al., 2023b) \cite{wang2023seggpt} by 15.4 and the original model by 0.3.

Furthermore, MN exhibits a substantial margin of improvement over PerSAM-F (Zhang et al., 2023) \cite{zhang2023personalize}, with an increase of +30.0 mean mIoU on COCO-20$^{i}$, +11.6 mIoU on FSS-1000 \cite{li2020fss}, and +21.7 mean mIoU on LVIS-92$^{i}$. Moreover, when evaluated on three different foundation models—DINOv2 \cite{oquab2023dinov2}, Resnet50 \cite{he2016deep}, and Efficientnet \cite{tan2019efficientnet}—MN demonstrates outstanding improvements in one-shot object part segmentation tasks. Specifically, MN outperforms the original models by an average of about 2 mean mIoU, showcasing the robust generality and flexibility of our method.
Additionally, our thorough investigation into drop ratios has bolstered the credibility and robustness of our NubbleDrop experiment.

Our main contributions can be summarized as follows:

(i) Through a rigorous mathematical discourse on feature interaction and fusion, as well as the uneven distribution of channel values, we have elucidated significant issues associated with directly computing the similarity of two patches using raw features extracted by VFMs. To validate our assertions, we conducted separate experiments, thereby laying the groundwork for future improvement studies.

(ii) We introduced a method that necessitates no training, incurs minimal computational demand, and demonstrated its ability not only to preserve the original model's capabilities but also to significantly address the aforementioned issues. Moreover, this method can be effortlessly implemented in various similarity computing scenarios.

(iii) We conducted extensive experiments to validate the effectiveness, robustness, and broad applicability of our approach. These experiments included assessing improvements achieved by NubbleDrop across different VFMs (such as ResNet, DINOv2, and Efficientnet), enhancements across various datasets, evaluation of performance at the image level, and exploration of the effects of discarding varying numbers of channels.

\section{Related Work}
\textbf{Feature Interactions:} Feature interactions refer to the contextual dependencies between features that collectively influence predictions. Various methods exist for extracting feature interactions in prediction models. Tsang et al. \cite{tsang2018detecting} introduce a novel framework for detecting statistical interactions captured by a feed-forward multi-layer neural network by directly interpreting its learned weights. Furthermore, Tsang et al. \cite{tsang2020does} propose an interaction attribution and detection framework called Archipelago, which offers more interpretable explanations for analyzing the impact of interactions on predictions, accompanied by visualizations of their approach.

\textbf{Vision Foundation Models:} Driven by extensive pre-training, foundational vision models have achieved remarkable success in computer vision. Drawing inspiration from the concept of masked language modeling \cite{devlin2018bert,liu2019roberta} in natural language processing, MAE \cite{he2022masked} adopts an asymmetric encoder-decoder architecture and implements masked image modeling to efficiently train scalable vision Transformer models \cite{dosovitskiy2020image}. CLIP \cite{radford2021learning} learns image representations from a vast corpus of 400 million image-text pairs, demonstrating impressive zero-shot image classification capabilities. Through image and patch-level discriminative self-supervised learning, DINOv2 \cite{oquab2023dinov2} acquires versatile visual features applicable to various downstream tasks. Recently, SAM \cite{kirillov2023segment}, pre-trained with 1 billion masks and 11 million images, has emerged with remarkable zero-shot, class-agnostic segmentation performance. Despite the exceptional performance of vision foundation models in fine-tuning, their capabilities remain limited in various visual perception tasks.

\textbf{Vision Generalist for Segmentation:} In recent times, there has been a growing endeavor to consolidate various segmentation tasks into a unified model leveraging the Transformer architecture \cite{vaswani2017attention}. The versatile Painter model \cite{wang2023images} reimagines the outcomes of diverse visual tasks as images and employs masked image modeling on continuous pixels for in-context training with labeled datasets. SegGPT \cite{wang2023seggpt}, a variant of the Painter model, introduces a novel random coloring method for in-context training to enhance the model's generalization capabilities. SEEM \cite{zou2024segment} effectively addresses various segmentation tasks by leveraging spatial queries such as points and textual prompts. More recently, PerSAM \cite{zhang2023personalize} extends SAM for personalized segmentation and video object segmentation with minimal training requirements, while Matcher \cite{liu2023matcher}, a training-free framework, endeavors to tackle various segmentation tasks in a single shot using all-purpose feature matching.

\section{Problem Analysis}
\subsection{Feature Interactions}
 A statistical interaction describes a situation in which the joint influence of multiple variables on an
 output variable is not additive (Dodge, 2006~\cite{dodge2003oxford}; Sorokina et al., 2008~\cite{sorokina2008detecting}). Let $x_{i}$, $i$ $\epsilon$ $[p]$ be the features
 and $y$ be the response variable, a statistical interaction $\mathbb{I}$ 
$\subseteq$ $[p]$ exists if and only if $\mathbb{E}$ $[y\mid x]$, which is
 a function of $\chi$ = ($x_{1}$, $x_{2}$,...,
 $x_{p}$), contains a non-additive interaction between variables $x_{\mathbb{I}}$:

\textbf{Definition} (Non-additive Interaction). Consider a function $f(\cdot)$ with input variables $x_{i}$, $i$ $\epsilon$ $[p]$, and an interaction $\mathbb{I}$ $\subseteq$ $[p]$. Then $\mathbb{I}$ is a non-additive interaction of function $f(\cdot)$ if and only if there does not exist a set of functions $f_{i}(\cdot)$, $\forall i$ $\epsilon$ $\mathbb{I}$ where $f_{i}(\cdot)$ is not a function of $x_{i}$, such that
\begin{equation}
f(\chi) = \sum_{i\epsilon\mathbb{I}}  f_{i} (\chi_{[p]\setminus\{i\}})
\end{equation}

\textbf{Interaction Strength:} In their work, Tsang et al. \cite{tsang2018detecting} demonstrate that the interaction strength $\omega_{i}(\mathbb{I})$ of a potential interaction $\mathbb{I}$ at the $i$-th unit in the first hidden layer $h_{i}^{(1)}$
\begin{equation}
    \omega_{i}(\mathbb{I}) = z_{i}^{(1)}\mu(|\boldsymbol{\mathbf{W}}_{i,\mathbb{I}}^{(1)}|)
\end{equation}

  where $\mu$(.) is the averaging function for an interaction that represents the interaction strength and the aggregated weight $z()$ is cumulative matrix multiplications of the absolute values of weight  matrices.
\begin{equation}
z^{(l)}= | \boldsymbol{\mathbf{w}}^{y} |^{\mathbf{T}}|\mathbf{W}^{(L)}| \cdot |\mathbf{W}^{(L-1)}|  \cdot\cdot\cdot |\mathbf{W}^{(l-1)}|  
\end{equation}

More details and aggregating strengths across hidden units can be read in supplementary materials.



Due to the complex feature interactions and fusion inherent in extracted features, it is evident that we can not deem that any single channel can adequately represent the uniqueness of a patch. In other words, certain channels may not be suitable for computing similarity.

\begin{figure}[ht]
    \centering
    \includegraphics[width=0.8\linewidth,height=6.5cm]{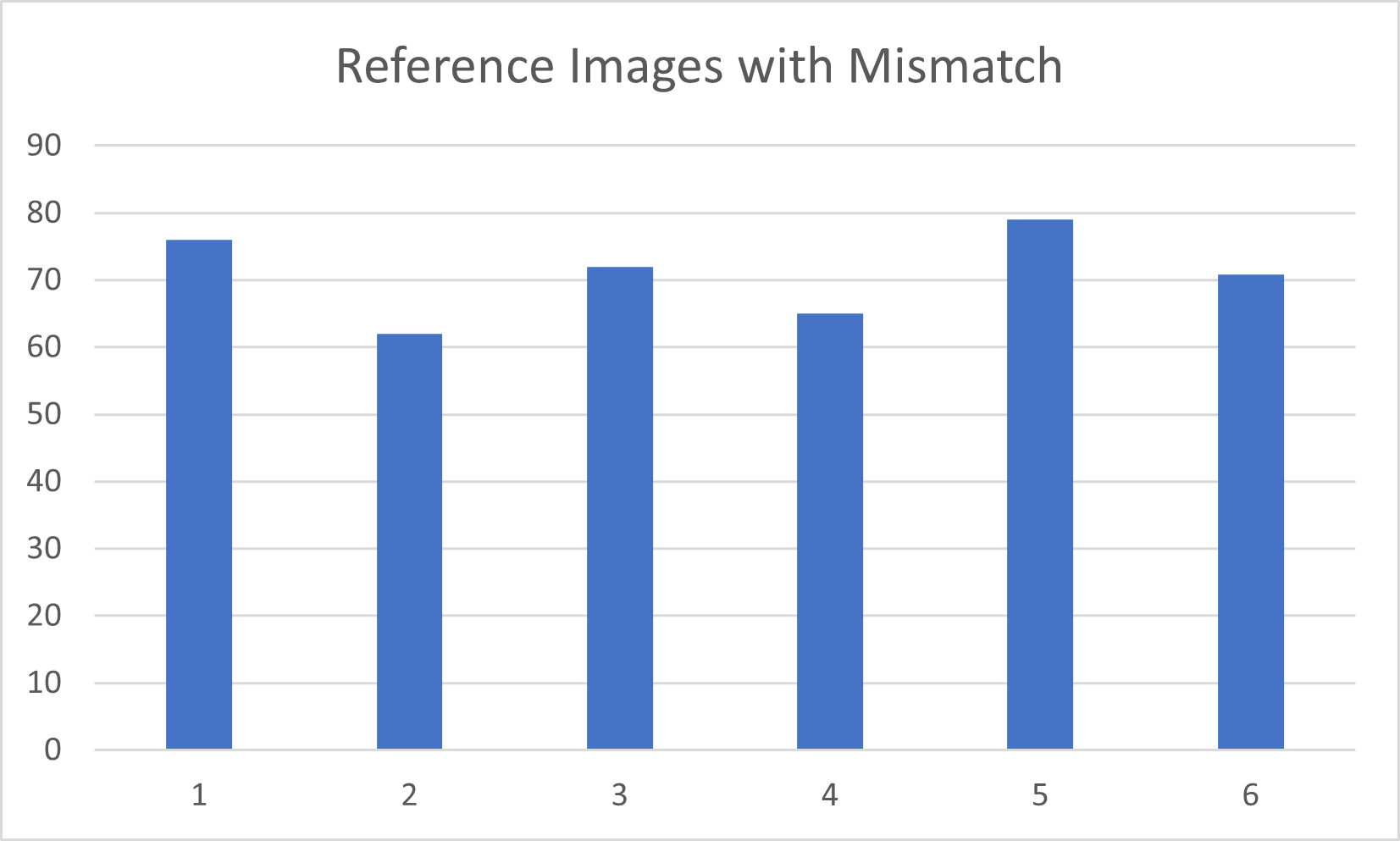}
    \caption{Numbers of reference images with mismatch in the randomly selected 500 pictures of COCO-20$^{i}$. On the x-axis, 1, 2, 3, 4, 5 represent dividing 500 into 5 parts for analysis, with 100 images each. 6 represents the average.}
    \label{fig:Figure 1}
\end{figure}

To illustrate our concern regarding this significant issue, we conducted an experiment on the COCO-20$^{i}$ dataset, wherein we computed the frequency of occurrences where not all target object points were best matched with foreground points (excluding themselves), as depicted in Figure \ref{fig:Figure 1}. Our findings reveal that in a considerable number of image tests, there is a substantial proportion of instances where incorrect matching occurs, amounting to approximately \textbf{70\%}. This outcome strongly validates our hypothesis and underscores the prevalence of mismatches when directly employing raw features for matching purposes.

\subsection{Uneven Distribution}



Upon normalization, the Euclidean norm ($L_{2}$ norm) of the tensor becomes 1. In theory, if the channel values are evenly distributed across each channel, a substantial amount of channel information will be effectively utilized. Conversely, if only a small number of channels dominate, they will exert a significant influence on computing cosine similarity, leaving the majority of channel information unused.

To validate this concern regarding uneven distribution in the feature channel dimension, we computed the extreme values and variances of channel values from features encoded across a large number of images. We define the phenomenon of "dominant channel" in the distribution of feature channel values when the maximum absolute channel value reaches 0.5 ($\kappa$) with a sum of squares greater than 0.25, representing a quarter of the total value. In this scenario, certain channel values are excessively large and play a crucial role in computing cosine similarity. Additionally, if the variance exceeds 0.0004 ($\nu$), it is deemed as the phenomenon of "channel submergence", wherein numerous channels possess values too small to be effectively utilized. Below is the mathematical representation:

\begin{equation}
    \mathcal{M}^{C} \ (\mid \mathcal{T}_{r} \mid) \ > \ \kappa
\end{equation}
\begin{equation}
    \mu(\ \mathcal{V}^{C} \ (\mid \mathcal{T}_{r} \mid) \ ) \ > \ \nu
\end{equation}



where $\mathcal{M}$ denotes the max value function and $\mu(\mathcal{V})$ represents the total average variance, while $C$ signifies the channel dimension. $\mathcal{T}_{r}$ denotes the feature tensors of reference images. It's worth noting that we also computed the total average of the mean values of the feature channels across sample images, yielding a value of 0.0230. If the total average variance exceeds 0.0004 ($\nu$), it indicates that, on average, each deviation surpasses 0.02, a value remarkably close to the mean. Hence, we can infer the existence of a "channel submergence" phenomenon, characterized by numerous channels with values too small to be effectively utilized.

\begin{figure}[ht]
    \centering
    \includegraphics[width=0.8\linewidth,height=6.5cm]{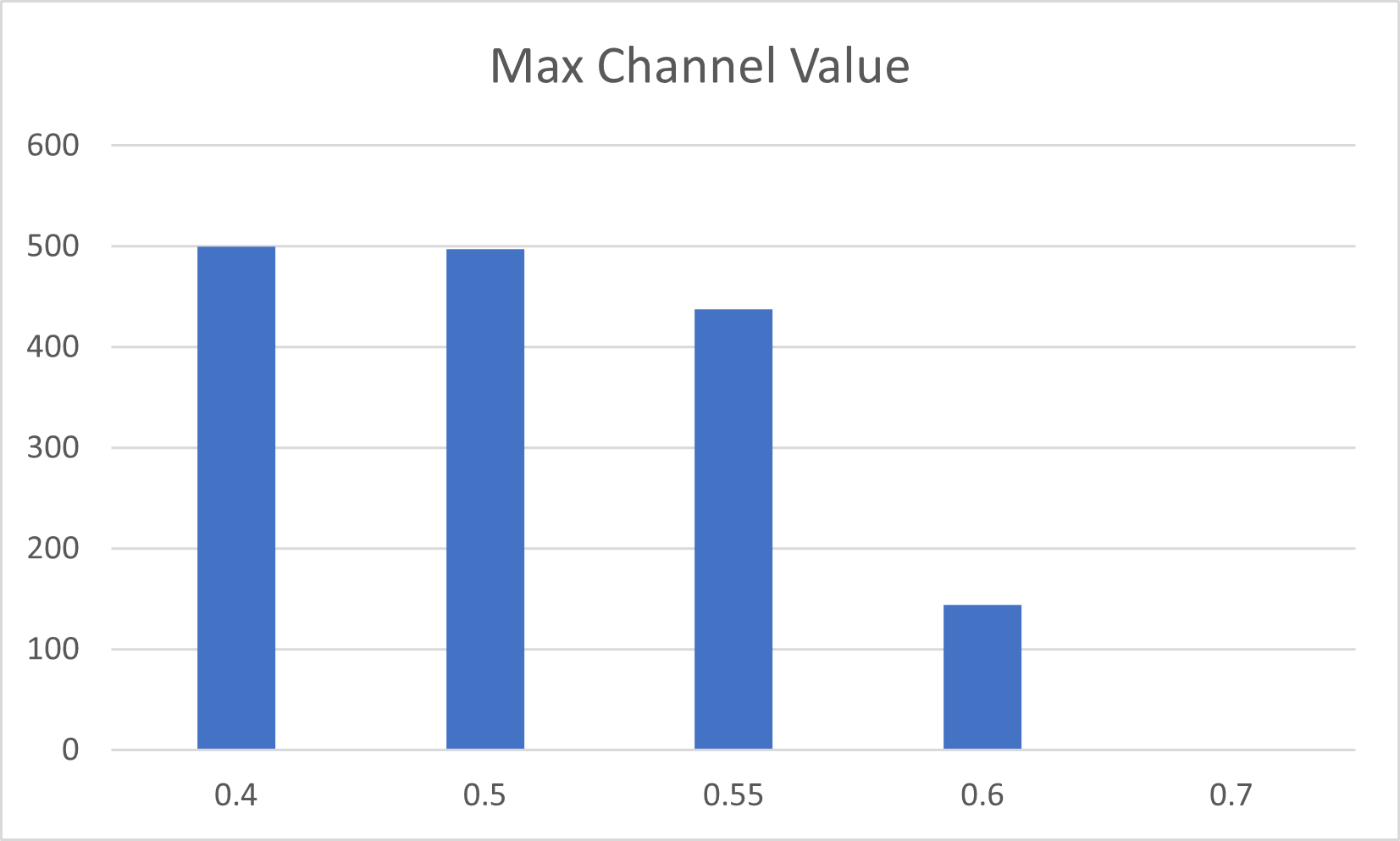}
    \caption{The y-axis represents the number of images from a random selection of 500 images from COCO-20$^i$ dataset, where the maximum channel value in the the normalized features encoded by DINOv2  exceeds the number indicated on the x-axis.}
    \label{fig:Figure 2}
\end{figure}

\begin{figure}[ht]
    \centering
    \includegraphics[width=0.8\linewidth,height=6.5cm]{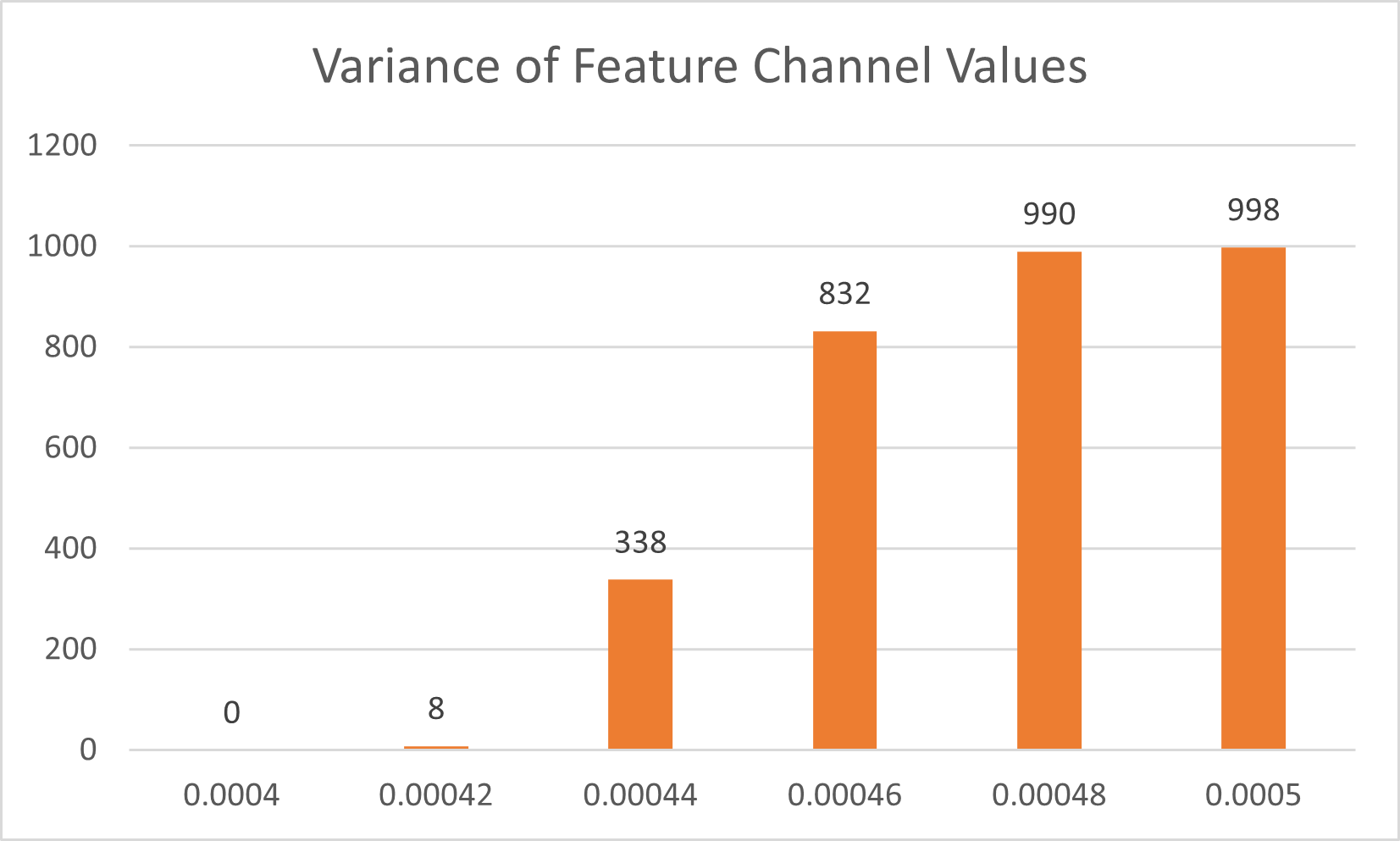}
    \caption{The y-axis represents the number of images from a random selection of 1000 images from COCO-20$^i$ dataset, where the variance of the the normalized features encoded by DINOv2  falls short of the number indicated on the x-axis. }
    \label{fig:Figure 3}
\end{figure}

The findings depicted in Figures \ref{fig:Figure 2} and \ref{fig:Figure 3} vividly illustrate the significant occurrences of "dominant channels" and "channel submergence" within the reference features.

\section{Methodology}
\subsection{Methods with Computational Lost}



To tackle the aforementioned problems, several methods can be explored:

Aligning features and removing inappropriate channels for computing similarity appears to be effective. However, since research on specific channels is still evolving, it's challenging to precisely determine the effects of each channel on every patch. Therefore, to implement this approach, we would need to calculate the similarity between a large number of target object patches and background patches. Then, iteratively removing inappropriate channels from each target patch would require a complexity close to $O((H\cdot W)^{2}\cdot K\cdot C)$, where $H$ and $W$ represent the height and width of the image patches, $K$ represents the number of compared background points per target object point, and $C$ represents the number of feature channels.

Additionally, trimming channels with extreme values is another strategy. To execute this idea, we would need to iterate through each channel value for every point, resulting in $(H\cdot W\cdot C)$ comparisons. It's essential to consider both the number of channels to trim and the cumulative sum of trimmed channel values, as they significantly influence model performance. However, the impact of all these extreme channel values on comparison remains unclear. Discarding them all while retaining other small channel values might have a counterproductive effect. To enhance the effectiveness and robustness of this method, extensive large-scale research is imperative.

\subsection{NubbleDrop with No Computation}

\begin{figure}[ht]
    \centering
    \includegraphics[width=1.01\linewidth,height=6.1cm]{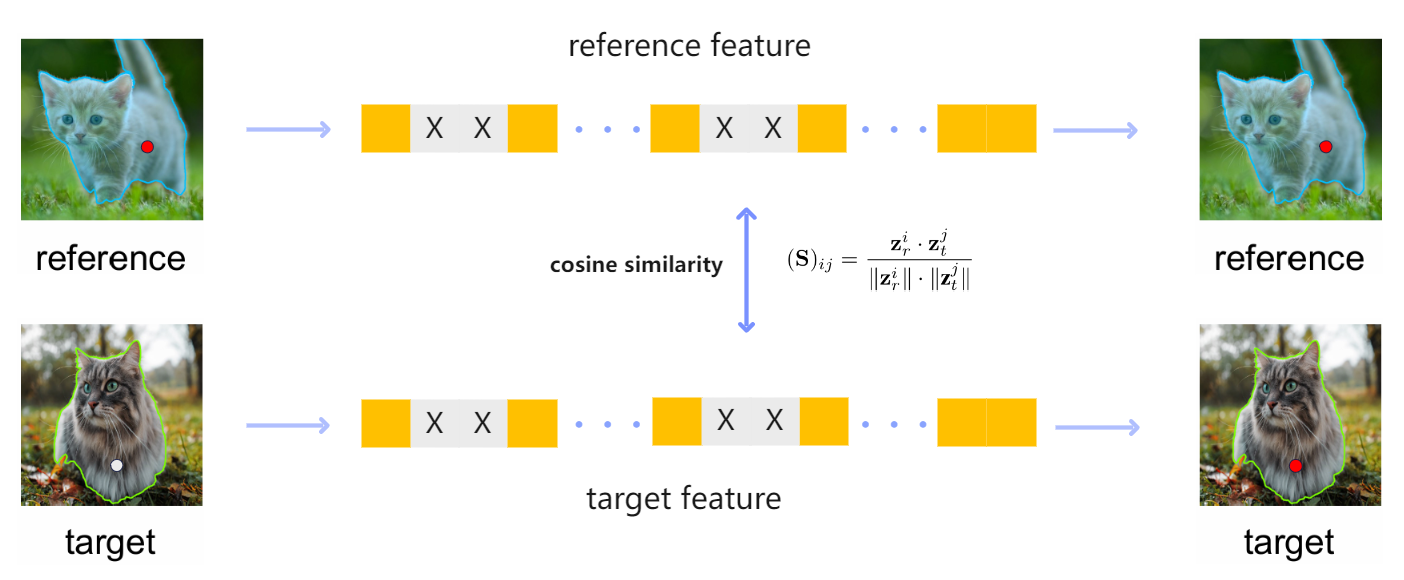}
    \caption{Illustration of NubbleDrop. Note that 'X' signifies that the channel is dropped, i.e., set to 0.}
    \label{fig:Figure 4}
\end{figure}

Traversing tensors suffers from poor parallelism, significantly impacting computation speed. Additionally, the lack of interpretive studies on specific feature channels introduces numerous considerations when targeting certain channel values for processing. To preserve the original model's exceptional performance while minimizing computational overhead and enhancing algorithm transferability and applicability, we propose NubbleDrop that randomly sets some channel values to zero.

Intuitively, if two features are very similar or very dissimilar, removing a small number of channels will likely have minimal impact on the original results. Therefore, NubbleDrop primarily affects ambiguous scenarios. As analyzed in Section 3, what influences the matching between patches are certain interaction channels or channels that play a decisive role. Since NubbleDrop randomly discards channels, if these problematic (Nubble) channels are retained, it's difficult to avoid poor matching results. However, if they are discarded, it can significantly improve the matching process.

Therefore, NubbleDrop can maintain the outstanding performance of the original model with a rather simple operation, while also potentially enhancing the model's performance to some extent. Furthermore, during a single test, matching between patch pairs is executed $H \cdot W$ times (where $H$ and $W$ represent the height and width of the image patches). Consequently, the model's performance typically improves when NubbleDrop is incorporated, as it's implemented over 1000 times within a single test.

The mathematical representation of NubbleDrop can be formalized as:
\begin{equation}
    R_i \ = \ \mathcal{E}(\mathcal{I}_r)
\end{equation}
\begin{equation}
    T_i \ = \ \mathcal{E}(\mathcal{I}_t)
\end{equation}
\begin{equation}
    k \ \sim \ sample(n)
\end{equation}
\begin{equation}
    R_i^j =  k_i^j * R_i^j ,\ \  T_i^j =  k_i^j * T_i^j
\end{equation}

where $R_{i}$ and $T_{i}$ represent the features extracted by VFMs ($\mathcal{E}$) of the reference image and target image, $k$ and formula (9) denote selecting a certain number of channels to set to 0 in the channel dimension with a total of $n$. 

\begin{figure}[ht]
    \centering
    \includegraphics[width=0.8\linewidth,height=6.5cm]{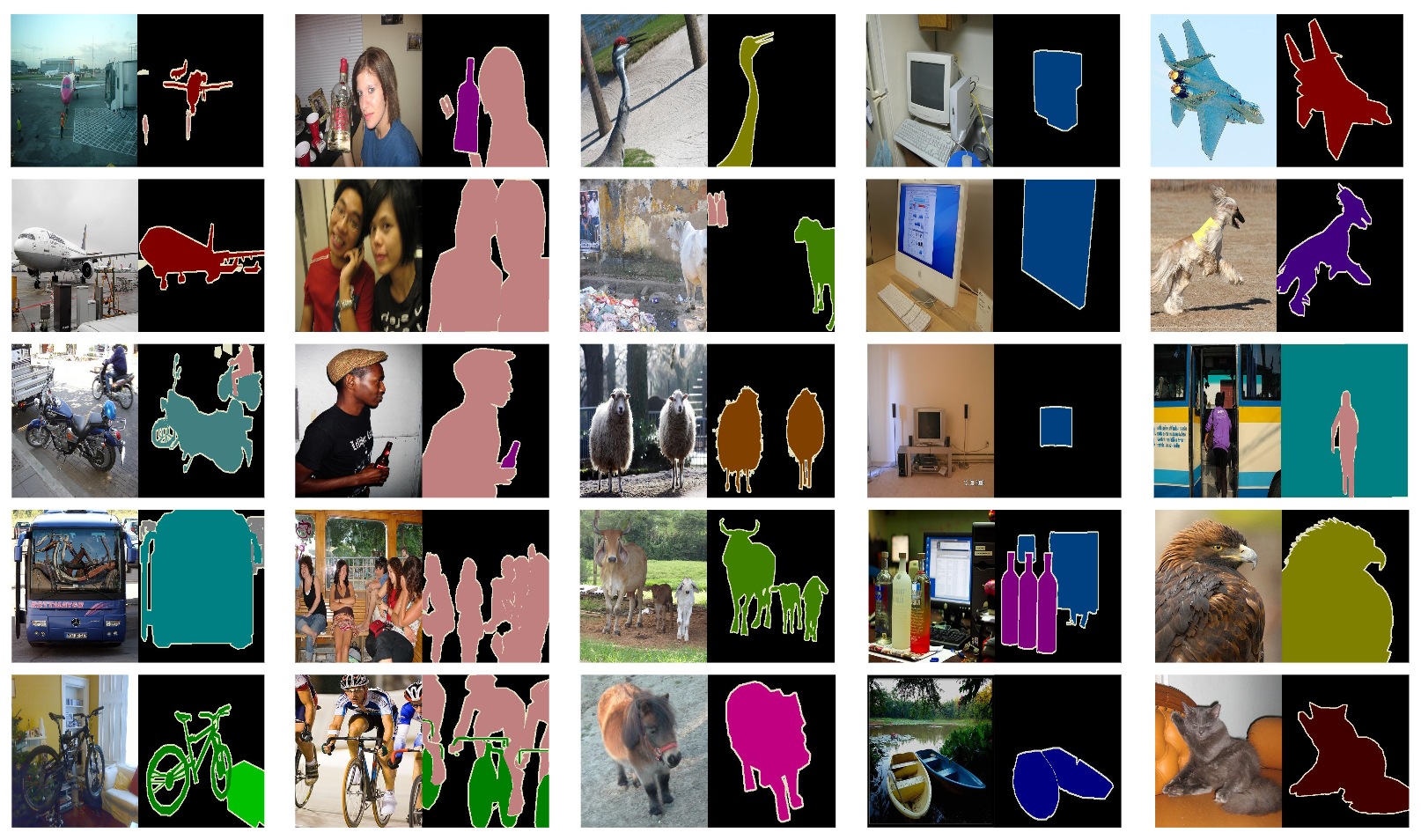}
    \caption{Examples of PASCAL-Part dataset. }
    \label{fig:Figure 5}
\end{figure}

 \begin{table}[]
\caption{Quantitative comparison with 8 methods on COCO-20$^{i}$, FSS-1000, and LVIS-92$^{i}$ dataset. \textcolor{gray}{Gray} indicates the model is trained by in-domain datasets.  Note that the training data of SegGPT includes COCO. $\pi$ indicates the training-free method.}
\centering
 \scalebox{1}[1]{
\begin{tabular}{ll|ll|l|l}
\multirow{2}{*}{Methods} & \multirow{2}{*}{Venue} & \multicolumn{2}{l|}{COCO-20i} & FSS-1000 & LVIS-92i \\
                         &                        & \multicolumn{2}{l|}{\makecell[c]{one-shot}} &\makecell[c]{ one-shot} & \makecell[c]{one-shot} \\ \hline
specialist model         &                        & \multicolumn{2}{l|}{}         &          &          \\
\textcolor{gray}{HSNet~\cite{min2021hypercorrelation}}                    & \textcolor{gray}{ICCV'21}                & \multicolumn{2}{l|}{\makecell[c]{\textcolor{gray}{41.2}}}     & \makecell[c]{\textcolor{gray}{86.5}}     & \makecell[c]{17.4}     \\
\textcolor{gray}{VAT~\cite{hong2022cost}}                      & \textcolor{gray}{ECCV'22}                & \multicolumn{2}{l|}{\makecell[c]{\textcolor{gray}{41.3}}}     & \makecell[c]{\textcolor{gray}{90.3}}     & \makecell[c]{{18.5}}     \\
\textcolor{gray}{FPTrans~\cite{zhang2022feature}}                  & \textcolor{gray}{NeurIPS'22}             & \multicolumn{2}{l|}{\makecell[c]{\textcolor{gray}{47.0}}}     & \makecell[c]{-}        & \makecell[c]{-}        \\
\textcolor{gray}{MSANet~\cite{iqbal2022msanet}}                   & \textcolor{gray}{arvix'22}               & \multicolumn{2}{l|}{\makecell[c]{\textcolor{gray}{51.1}}}     & \makecell[c]{-}        & \makecell[c]{-}        \\ \hline
generalist model         &                        & \multicolumn{2}{l|}{}         &          &          \\
\textcolor{gray}{Painter~\cite{wang2023images}}                  & \textcolor{gray}{CVPR'23}                & \multicolumn{2}{l|}{\makecell[c]{\textcolor{gray}{33.1}}}     & \makecell[c]{61.7}     & \makecell[c]{10.5}     \\
\textcolor{gray}{SegGPT~\cite{wang2023seggpt}}                   & \textcolor{gray}{ICCV'23}                & \multicolumn{2}{l|}{\makecell[c]{\textcolor{gray}{56.1}}}     & \makecell[c]{85.6}     & \makecell[c]{18.6}     \\
PerSAM-F$^{\pi}$~\cite{zhang2023personalize}                 & ICLR'24                & \multicolumn{2}{l|}{\makecell[c]{23.5}}     & \makecell[c]{75.6}     & \makecell[c]{12.3}     \\
Matcher$^{\pi}$~\cite{liu2023matcher}                  & ICLR'24                & \multicolumn{2}{l|}{\makecell[c]{52.1}}     & \makecell[c]{87.0}     & \makecell[c]{33.7}     \\
Matcher+NubbleDrop$^{\pi}$       & this work              & \multicolumn{2}{l|}{\makecell[c]{\textbf{53.5}}}     & \makecell[c]{\textbf{87.2}}     & \makecell[c]{\textbf{34.0}}    
\end{tabular}
}
\end{table}

\begin{table}[h]
\centering
\label{Table 2}
\caption{Quantitative comparison with 6 methods on PASCAL-part dataset. Notes that $\pi$ indicates the training-free method.}
\begin{tabular}{cc|clccl}
\multirow{2}{*}{Methods} & \multirow{2}{*}{Venue} & \multicolumn{5}{c}{PASCAL-Part}                                                                         \\
                         &                        & \multicolumn{1}{l}{animals} & indoor & \multicolumn{1}{l}{person} & \multicolumn{1}{l}{vehicles} & mean \\ \hline
HSNet~\cite{min2021hypercorrelation}                    & ICCV'21                & 21.2                        & 53.0   & 20.2                       & \multicolumn{1}{c|}{35.1}    & 32.4 \\
VAT~\cite{hong2022cost}                      & ECCV'22                & 21.5                        & 55.9   & 20.7                       & \multicolumn{1}{c|}{36.1}    & 33.6 \\
Painter~\cite{wang2023images}                  & CVPR'23                & 20.2                        & 49.5   & 17.6                       & \multicolumn{1}{c|}{34.4}    & 30.4 \\
SegGPT ~\cite{wang2023seggpt}                  & ICCV'23                & 22.8                        & 50.9   & 31.3                       & \multicolumn{1}{c|}{38.0}    & 35.8 \\
PerSAM$^{\pi}$ ~\cite{zhang2023personalize}                  & ICLR'24                & 19.9                        & 51.8   & 18.6                       & \multicolumn{1}{c|}{32.0}    & 30.1 \\
Matcher$^{\pi}$  ~\cite{liu2023matcher}                & ICLR'24                & 37.1                        & 56.3   & 32.4                       & \multicolumn{1}{c|}{45.7}    & 42.9 \\
Matcher+NubbleDrop$^{\pi}$      & this work              & 37.1                        & 56.2   & 32.2                       & \multicolumn{1}{c|}{45.6}    & 42.8
\end{tabular}
\end{table}

\begin{table}[]
\caption{Quantitative comparison with 3 different vision foundation models on COCO-20i dataset. Notes "-" means without NubbleDrop, "+" means adding NubbleDrop.}
\centering
\label{Table 3}
\scalebox{1.5}[1.5]{
\begin{tabular}{|c|c|cl|}
\hline
\multirow{2}{*}{Methods} & \multirow{2}{*}{VFM} & \multicolumn{2}{c|}{COCO-20i}            \\ \cline{3-4} 
                         &                      & \multicolumn{1}{l|}{\makecell[c]{-}}     & \makecell[c]{ +} \\ \hline
\multirow{3}{*}{Matcher} & DINOv2               & \multicolumn{1}{c|}{\makecell[c]{52.1}}  &\makecell[c]{ 53.5}        \\
                         & Resnet50             & \multicolumn{1}{c|}{\makecell[c]{18.59}} & \makecell[c]{22.75}       \\
                         & Efficientnet      & \multicolumn{1}{c|}{\makecell[c]{23.36}} & \makecell[c]{23.76}       \\ \hline
\end{tabular}
}
\end{table}

\section{Experiments}
\subsection{Datasets}

We assess the performance of Matcher with \textbf{NubbleDrop} across three benchmark datasets: COCO-20$^i$ \cite{nguyen2019feature}, FSS-1000 \cite{li2020fss}, and LVIS-92$^i$ \cite{liu2023matcher}. 

COCO-20$^i$ divides the 80 categories of the MSCOCO dataset \cite{lin2014microsoft} into four cross-validation folds, each comprising 60 training classes and 20 test classes. FSS-1000 comprises mask-annotated images from 1,000 classes, with 520, 240, and 240 classes in the training, validation, and test sets, respectively. LVIS-92$^i$ is a more challenging benchmark with a total of 920 classes, divided into 10 equal folds for testing purposes.

Additionally, we consider PASCAL-Part \cite{liu2023matcher}, a one-shot part segmentation dataset based on PASCAL VOC2010 and its body part annotations. This dataset consists of four superclasses—animals, indoor, person, and vehicles—with a total of 56 different object parts.
 
\subsection{Results}
We compare Matcher with the addition of NubbleDrop (\textbf{MN}) against a range of specialist models, including HSNet \cite{min2021hypercorrelation}, VAT \cite{hong2022cost}, FPTrans \cite{zhang2022feature}, and MSANet \cite{iqbal2022msanet}, as well as generalist models such as Painter \cite{wang2023images}, SegGPT \cite{wang2023seggpt}, PerSAM \cite{zhang2023personalize}, and Matcher \cite{liu2023matcher}. 

As illustrated in Table 1, MN achieves a mean mIoU of 53.5\% on COCO-20$^i$, with a 1.4\% enhancement over the original model, surpassing the state-of-the-art specialist model MSANet and performing comparably with SegGPT. It's worth noting that SegGPT's training data includes COCO. For FSS-1000, MN demonstrates highly competitive performance compared to specialist models and outperforms all generalist models without any performance degradation compared to the original model. 

In the case of LVIS-92$^i$, we evaluate the cross-dataset generalization abilities of MN and other models. MN achieves a mean mIoU of 34.0\%, surpassing the state-of-the-art generalist model SegGPT by 15.4\% and enhancing Matcher by 0.3 mean mIoU. These results underscore the remarkable robustness of NubbleDrop across different datasets.


Furthermore, we conduct experiments with NubbleDrop on three distinct vision foundation models, revealing that MN significantly outperforms those models utilizing raw features. Specifically, MN exhibits enhancements of 1.4, 4.16, and 0.4 mean mIoU in DINOv2, Resnet50, and Efficientnet, respectively, without any degradation in performance. These results underscore the remarkable flexibility and transferability of our method.

\subsection{Further Experimentation}

We've conducted two additional experiments and obtained two result graphs. One is a line graph depicting the average performance improvement as the number of test images increases. The other graph illustrates the impact of drop ratio variations on model performance. 

The results of the first experiment indicate that the improvement achieved by our method with a small number of images is consistent with the overall improvement, demonstrating the effectiveness of our method on individual images and its ability to preserve the performance of the original model.

\begin{figure}[ht]
    \centering
    \includegraphics[width=0.8\linewidth,height=5.6cm]{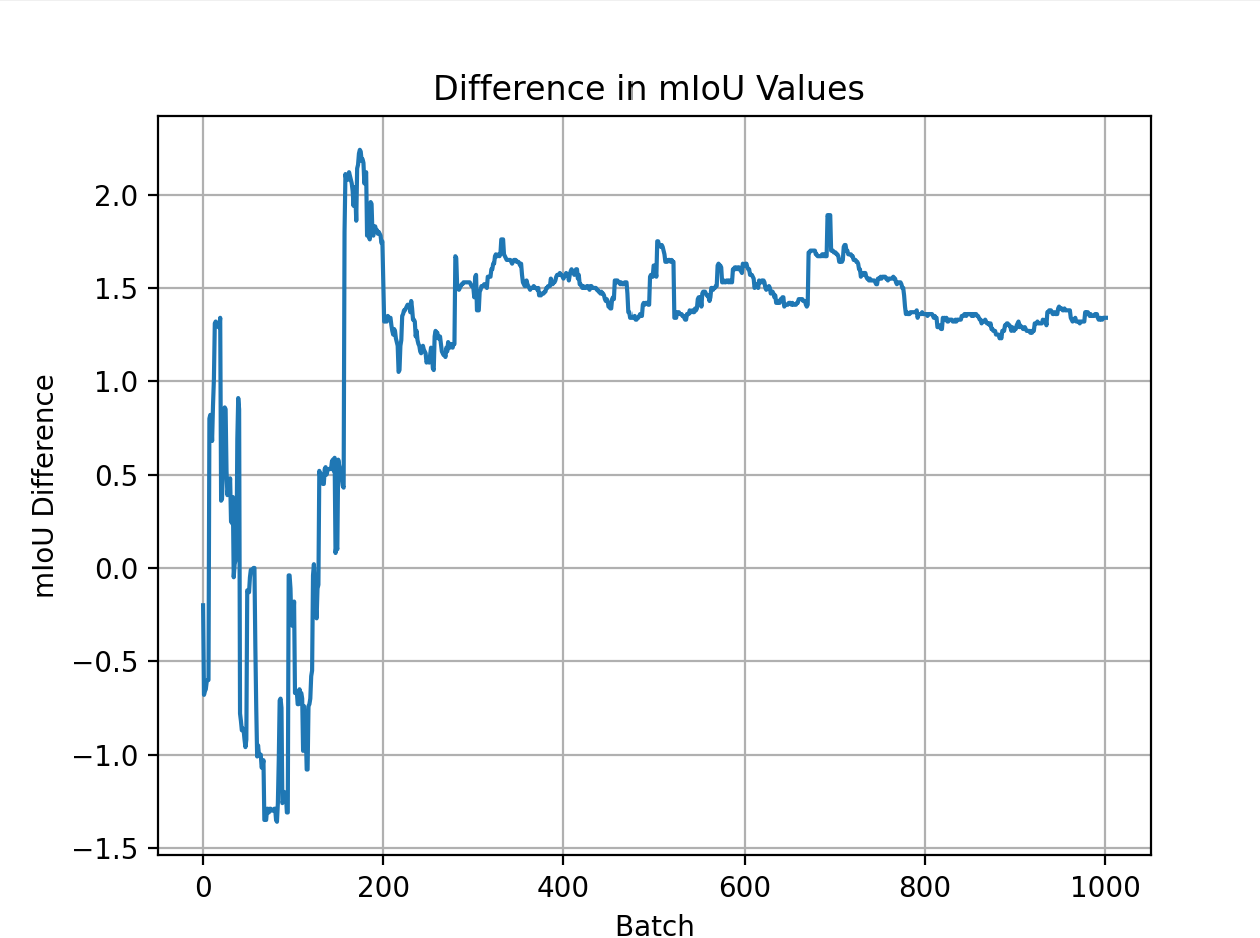}
    \caption{The x-axis represents the increase in the number of test images, while the y-axis represents the average improvement (mIoU).}
    \label{fig:Figure 5}
\end{figure}

\begin{figure}[ht]
    \centering
    \includegraphics[width=0.8\linewidth,height=5.6cm]{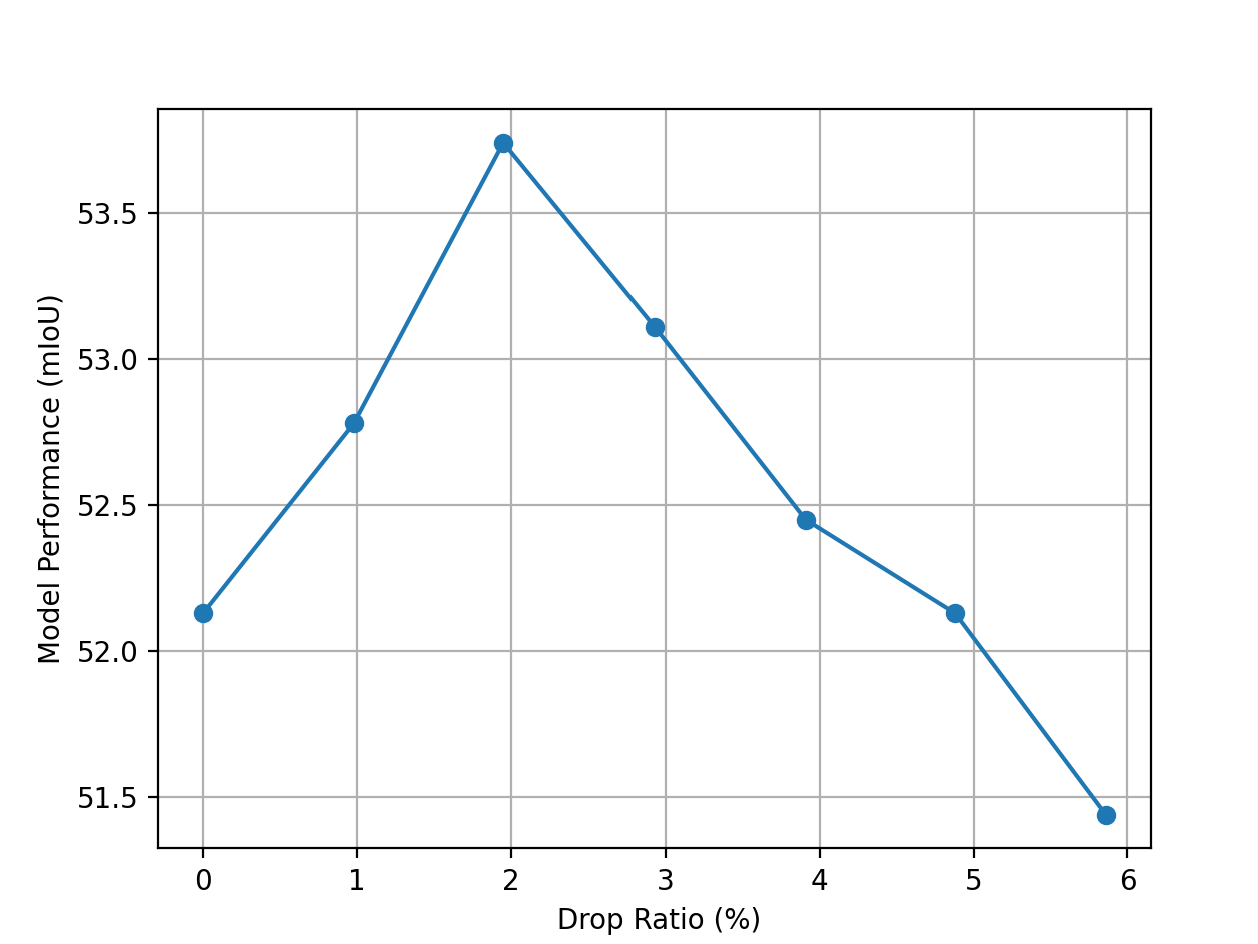}
    \caption{The x-axis represents the drop ratio of feature channels , while the y-axis represents the model performance (mIoU).}
    \label{fig:Figure 6}
\end{figure}

\section{Conclusion}
In this paper, we first analyze two existing problems in matching strategies and propose an exceptionally simple solution, NubbleDrop, which requires no training and minimal computational resources to address these issues. We also explore alternative methods that entail computational costs and conduct extensive experiments to validate the efficacy, scalability, and effectiveness of our approach.

\textbf{Limitations and Future Research}. Although our method is rather simple and does not introduce any computational overhead, the improvement in performance is not very significant. We have analyzed two problems we identify in existing matching strategies, providing useful directions for future research. We believe that there will be more effective methods to address these issues in the future.

{\small
\nocite{*}
\bibliographystyle{ieee_fullname}
\bibliography{neurips_2024}
}

\end{document}